\documentclass[runningheads]{llncs}
\usepackage{graphicx}
\usepackage{comment}
\usepackage{amsmath,amssymb} % define this before the line numbering.
\usepackage{color}

\newcommand{\etal}{\textit{et al}.}

\begin{document}
% \linenumbers
\pagestyle{headings}
\mainmatter
\def\ECCVSubNumber{5609}  % Insert your submission number here

\title{Toward Accurate and Realistic Virtual Try-on Through Shape Matching and Multiple Warps} % Replace with your title

\begin{comment}
\titlerunning{ECCV-20 submission ID \ECCVSubNumber} 
\authorrunning{ECCV-20 submission ID \ECCVSubNumber} 
\author{Anonymous ECCV submission}
\institute{Paper ID \ECCVSubNumber}
\end{comment}
%******************

%\begin{comment}
\titlerunning{Virtual Try-on Through Shape Matching and Multiple Warps}
\author{Kedan Li\inst{1} \and
Min Jin Chong\inst{1} \and
Jingen Liu\inst{2} \and David Forsyth\inst{1}}
\authorrunning{K. Li et al.}
% If there are more than two authors, 'tal' is used.
%
\institute{University of Illinois at Urbana-Champaign\\
\email{\{kedanli2,mchong6,daf\}@illinois.edu}\\
\and
JD AI Research\\
\email{jingen.liu@jd.com}}
%\end{comment}
%******************
\maketitle

\begin{abstract}
A virtual try-on method takes a product image and an image of a model and produces an image of the model wearing the product. Most methods essentially compute warps from the product image to the model image and combine using image generation methods.  However, obtaining a realistic image is challenging because the kinematics of garments is complex and because outline, texture, and shading cues in the image reveal errors to human viewers.  The garment must have appropriate drapes; texture must be warped to be consistent with the shape of a draped garment;  small details (buttons, collars, lapels, pockets, etc.) must be placed appropriately on the garment, and so on. Evaluation is particularly difficult and is usually qualitative.

This paper uses quantitative evaluation on a challenging, novel dataset to demonstrate that (a) for any warping method, one can choose target models automatically to improve results, and (b) learning multiple coordinated specialized warpers offers further improvements on results. Target models are chosen by a learned embedding procedure that predicts a representation of the products the model is wearing.  This prediction is used to match products to models. Specialized warpers are trained by a method that encourages a second warper to perform well in locations where the first works poorly. The warps are then combined using a U-Net. Qualitative evaluation confirms that these improvements are wholesale over outline, texture shading, and garment details.
\end{abstract}

\section{Introduction}

E-commerce means not being able to try on a product, which is difficult for fashion consumers~\cite{Vaccaro:2018:DFP:3173574.3174201}.  Sites now routinely put up photoshoots of models wearing products,
but volume and turnover mean doing so is very expensive and time consuming~\cite{fashionindustry2019}. There is a need to generate realistic and accurate images of fashion models wearing different sets of clothing. 
One could use 3D models of posture~\cite{Deep3DPose,drape}.  The alternative -- synthesize product-model images without 3D measurements~\cite{han2017viton,wang2018toward,Rocco17,Dong2018SoftGatedWF,Han_2019_ICCV} -- is known as virtual try-on. 
These methods usually consist of two components: 1) a spatial transformer to warp the product image using some estimate of the model's pose and 2) an image generation network that combines the coarsely aligned,
warped product with the model image to produce a realistic image of the model wearing the product.

\begin{figure}[t!]
\begin{center}
   \includegraphics[width=0.8\linewidth]{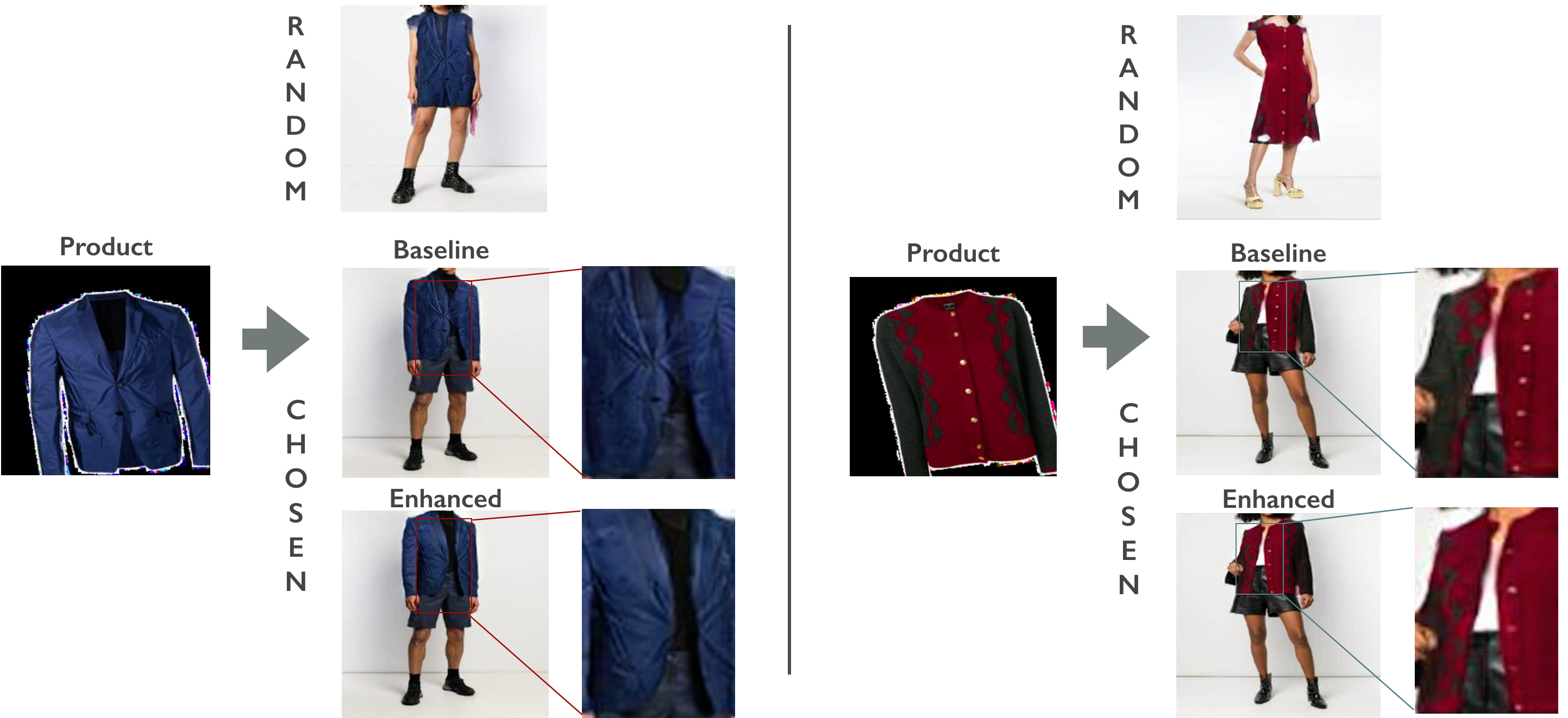}
\end{center}
   \caption{
   		Translating a product to a poorly chosen model leads to difficulties (random model; notice how the blazer has been squashed on
the {\bf left}, and the jersey stretched on the {\bf right}). Our method can choose a good target model for a given product, leading to significant qualitative and quantitative improvement in transfers (chosen model). In addition, we train multiple warpers to act in a coordinated fashion, which further enhances the generation results (enhanced; the buttonholes on the jacket are in the right place {\bf left}, and the row of buttons on
the cardigan is plausible {\bf right}). The figure shows that (a) carefully choosing the model to warp and (b) using multiple specialized warpers significantly improve the transfer. Quantitative results in table~\ref{table:fid} strongly supports the two points made.
\vspace{-0.5cm}
   		}
\label{fig:overview}
\end{figure}

It is much easier to transfer with simple garments like t-shirts, which are emphasized in the literature. General garments (unlike t-shirts) might open in front; have sophisticated drapes; have shaped structures like collars and cuffs; have buttons; and so on.  These effects severely challenge existing methods (examples in Supplementary Materials).   Warping is significantly improved if one uses the product image to choose a model image that is suited to that garment (Figure~\ref{fig:overview}). 

\begin{figure}
\begin{center}
   \includegraphics[width=0.8\linewidth]{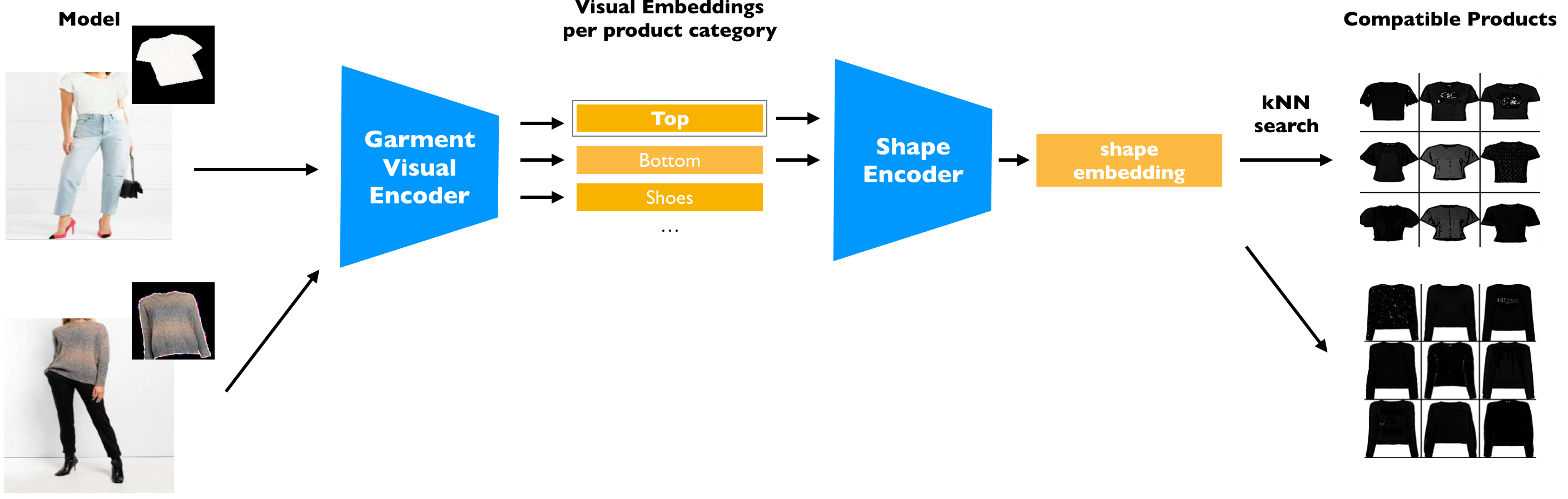}
\end{center}
   \caption{
   		It is hard to transfer, say, a long sleeved shirt onto a model wearing a t-shirt. Our process retrieves compatible pairs in two stages. First, we compute a garment appearance embedding using a garment visual encoder, trained using product-model pairs and spatial attention. Then, a shape encoder computes the shape embedding from the garment appearance embedding. The shape embedding is learned using product contour as metric, which only preserves shape information. When we transfer, we choose a model wearing a compatible garment by searching in the shape embedding space.
\vspace{-0.5cm}
}
\label{fig:shape-embeddinng-simple}
\end{figure}

At least in part, this is a result of how image generation networks are trained.  We train using paired images -- a product and a model wearing a product~\cite{han2017viton,wang2018toward,YuVTNFPAI}. This means that the generation network
always expects the target image to be appropriate for the product (so it is not trained to, for example, put a sweater onto a model wearing a dress, Figure~\ref{fig:overview}).   An alternative is to use adversarial training~\cite{Dong2018SoftGatedWF,Dong2019TowardsMG,Raj2018SwapNetIB,Grigorev2019CoordinateBasedTI,Raffiee2020GarmentGANPA}; but it is hard to preserve specific product details (for example, a particular style of buttons; a decal on a t-shirt) in this framework. To deal with this difficulty, 
we learn an embedding space for choosing product-model pairs that will result in high-quality transfers (Figure~\ref{fig:shape-embeddinng-simple}).
The embedding learns to predict what shape a garment in a model image would take if it were in a product image. Products are then matched to models wearing similarly shaped garments.
Because models typically wear many garments, we use a spatial attention visual encoder to parse each category (top, bottom, outerwear, all-body, etc.) of garment and embed each separately.

Another problem arises when a garment is open (for example, an unbuttoned coat).  In this case, the target of the warp might have
more than one connected component.  Warpers tend to react by fitting one region well and the other poorly, resulting in misaligned details (the buttons of Figure~\ref{fig:overview}). 
Such errors may make little contribution to the training loss, but are very apparent and are considered severe problems by real users.
We show that using multiple coordinated specialized warps produces substantial quantitative and qualitative improvements in warping. 
Our warper produces multiple warps, trained to coordinate with each other. An inpainting network combines the warps and the masked model, and creates a synthesized image. The inpainting network essentially learns to choose between the warps, while also provides guidance to the warper, as they are trained jointly. Qualitative evaluation confirms that an important part of the improvement results from better predictions of buttons, pockets, labels, and the like.

We show large scale quantitative evaluations of virtual try-on.  We collected a new dataset of 422,756 pairs of product images and studio photos by mining fashion e-commerce sites. The dataset contains multiple product categories. We compare with prior work on the established VITON dataset~\cite{han2017viton} both quantitatively and qualitatively. Quantitative result shows that choosing the product-model pairs using our shape embedding yields significant improvements for all image generation pipelines (table~\ref{table:fid}). Using multiple warps also consistently outperform the single warp baseline, demonstrated through both quantitative (table~\ref{table:fid}, figure \ref{fig:comparison-K}) and qualitative (figure~\ref{fig:multi-warp-qualitative}) results. Qualitative comparison with prior work shows that our system preserves the details of both the to-change garment and the target model more accurately than prior work. We conducted a user study simulating the cost for e-commerce to replace real model with synthesized model. Result shows 40\% of our synthesized model are thought as real models.

As a summary of our contributions:
\begin{itemize}
	\item we introduce a matching procedure that results in significant qualitative and quantitative improvements in virtual try-on, whatever warper is used.
	\item we introduce a warping model that learns multiple coordinated-warps and consistently outperforms baselines on all test sets.
	\item our generated results preserve details accurately and realistically enough to make shoppers think that some of the synthesized images are real. 
\end{itemize}

\section{Related Work}

{\bf Image synthesis:} Spatial transformer networks estimate geometric transformations using neural networks~\cite{NIPS2015_5854}. 
Subsequent work~\cite{warpnet,Rocco17} shows how to warp one object to another. Warping can be used to produce images of rigid objects~\cite{Ji2017DeepVM,lin2018stgan} and non-rigid objects (e.g., clothing)~\cite{han2017viton,Dong2019TowardsMG,wang2018toward}. 
In contrast to prior work, we use multiple spatial warpers.  

Our warps must be combined into a single image, and our U-Net for producing this image follows trends in inpainting (methods that fill in missing portions of an image, see~\cite{Yang_2017_CVPR,Liu_2018_ECCV,Yu2018GenerativeII,Yu_2019_ICCV}).  Han \etal~\cite{Han2019CompatibleAD,inpainting-based} show inpainting methods can complete missing clothing items on people. 

In our work, we use FID$_\infty$ to quantitatively evaluate our method. This is based on the Fréchet Inception Distance (FID)~\cite{heusel2017gans},  a common metric in generative image modelling~\cite{brock2018large,zhang2018self,karras2019style}. Chong \etal ~\cite{chong2019effectively} recently showed that FID is biased; extrapolation removes the bias,
to an unbiased score (FID$_\infty$).

{\bf Generating clothed people:} Zhu \etal~\cite{zhu2017be} used a conditional GAN to generate images based on pose skeleton and text descriptions of garment. SwapNet~\cite{Raj2018SwapNetIB} learns to transfer clothes from person A to person B by disentangling clothing and pose features. Hsiao \etal~\cite{hsiao2019fashionplus} learned a fashion model synthesis network using per-garment encodings to enable convenient minimal edit to specific items.  In contrast, we warp products onto real model images.

{\bf Shape matching} underlies our method to match product to model.  Tsiao \etal~\cite{Hsiao2019DressingFD} built a shape embedding to enable matching between human body and well-fitting clothing items. 
Prior work estimated the shape of human body~\cite{Bogo:ECCV:2016,Kanazawa2017EndtoEndRO}, clothing items~\cite{Danerek2017DeepGarment3,Jeong2015GarmentCF} and both~\cite{Natsume2019SiCloPeS,Saito2019PIFuPI}, through 2D images.  The DensePose~\cite{Guler_2018_CVPR} descriptor helps modeling the deformation and shading of cloth and, therefore, has been adopted by recent work~\cite{Neverova2018DensePT,Grigorev2019CoordinateBasedTI,Wu2018M2ETryON,8836494,Chen_2019_ICCV_Workshops,inpainting-based}. 

{\bf Virtual try-on} (VTO) maps a product to a model image.  VITON~\cite{han2017viton} uses a U-Net to generate a coarse synthesis and a mask on the model where the product is presented. A mapping from the product mask to the on-model mask is learned through Thin plate spline (TPS) transformation~\cite{993558}. The learned mapping is applied on the product image to create a warp. Following their work, Wang \etal~\cite{wang2018toward} improved the architecture using a Geometric Matching Module~\cite{Rocco17} to estimate the TPS transformations parameters directly from pairs of product image and target person. They train a separate refinement network to combine the warp and the target image. VTNFP~\cite{YuVTNFPAI} extends the work by incorporatiing body segments prediction and later works follow similar procedure~\cite{Raffiee2020GarmentGANPA,Jandial2020SieveNetAU,Song2019SPVITONSI,Lee_2019_ICCV_Workshops,Ayush_2019_ICCV_Workshops}. However, TPS transformation fails to produce reasonable warps, due to the noisiness of generated masks in our dataset, as shown in Figure~\ref{fig:comparison-VITON} right. Instead, we adopt affine transformations which we have found to be more robust to imperfections instead of TPS transformation.
A group of following work extended the task to multi-pose. Warping-GAN~\cite{Dong2018SoftGatedWF} combined adversarial training with GMM, and generate post and texture separately using a two stage network. MG-VTON~\cite{Dong2019TowardsMG} further refine the generation method using a three-stage generation network. Other work~\cite{Hsieh2019FashionOnSI,arbitrary-pose,8836494,Chen_2019_ICCV_Workshops,Wang2019DownTT} followed similar procedure. Han \etal~\cite{Han_2019_ICCV} argued that TPS transformation has low degree of freedom and proposed a flow-based method to create the warp.

Much existing virtual try-on work ~\cite{han2017viton,Dong2019TowardsMG,Hsieh2019FashionOnSI,Wu2018M2ETryON,arbitrary-pose,YuVTNFPAI,Jandial2020SieveNetAU,Raffiee2020GarmentGANPA} is evaluated on datasets that only have tops (t-shirt, shirt, etc.). Having only tops largely reduces the likelihood of shape mismatch as tops have simple and similar shapes. In our work, we extend the problem to include clothing items of all categories(t-shirt, shirt, pants, shorts, dress, skirt, robe, jacket, coat, etc.), and propose a method for matching the shape between the source product and the target model. Evaluation shows that using pairs that match in shape significantly increases the generation quality for both our and prior work (table~\ref{table:fid}).

In addition, real studio outfits are often covered by an unzipped/unbuttoned outerwear, which is also not presented in prior work~\cite{han2017viton,Dong2019TowardsMG,Hsieh2019FashionOnSI,Wu2018M2ETryON,arbitrary-pose,YuVTNFPAI,Raffiee2020GarmentGANPA}. This can cause partition or severe occlusion to the garment, and is not addressed by prior  work as shown in Figure~\ref{fig:comparison-VITON}. We show that our multi-warp generation module ameliorates these difficulties.

\section{Proposed Method}

\begin{figure}
\begin{center}
   \includegraphics[width=0.7\linewidth]{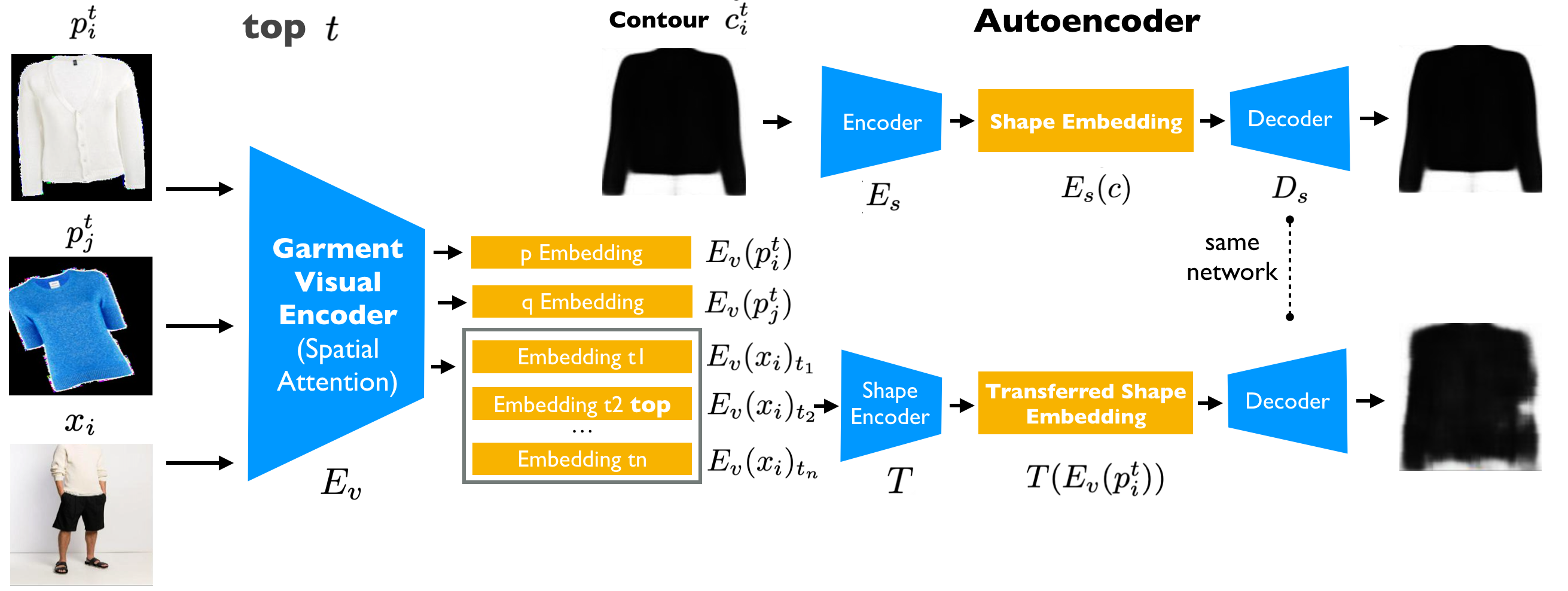}
\end{center}
   \caption{The shape matching net (SMN) is trained using triplets of: a model $x_i$ wearing a product $p_i^t$ of type $t$ (e.g., `top');  that product $p_i^t$; and a distractor product $p_j^t, j \neq i$. An autoencoder is trained to produce codes representing the contour mask $c_i^t$ of $p_i^t$. The visual encoder
must then produce a representation of $x_i$ from each type $t$ that lies close to $p_i^t$'s representation and far from $p_j^t$'s representation -- roughly, the encoder tries to make a representation of the frontal appearance of the product. This representation is passed through a shape encoder to produce a code, and $p_i^t$'s code must produce a reconstruction of $p_i^t$'s contour when passed through the autoencoder's decoder. In turn, this means that the shape embedding encodes the contour
of the {\em product image} of the product the model is wearing, so that the model can be matched to other such products.
\vspace{-0.5cm}
}
\label{fig:shape-embedding}
\end{figure}

\begin{figure}
\begin{center}
   \includegraphics[width=0.7\linewidth]{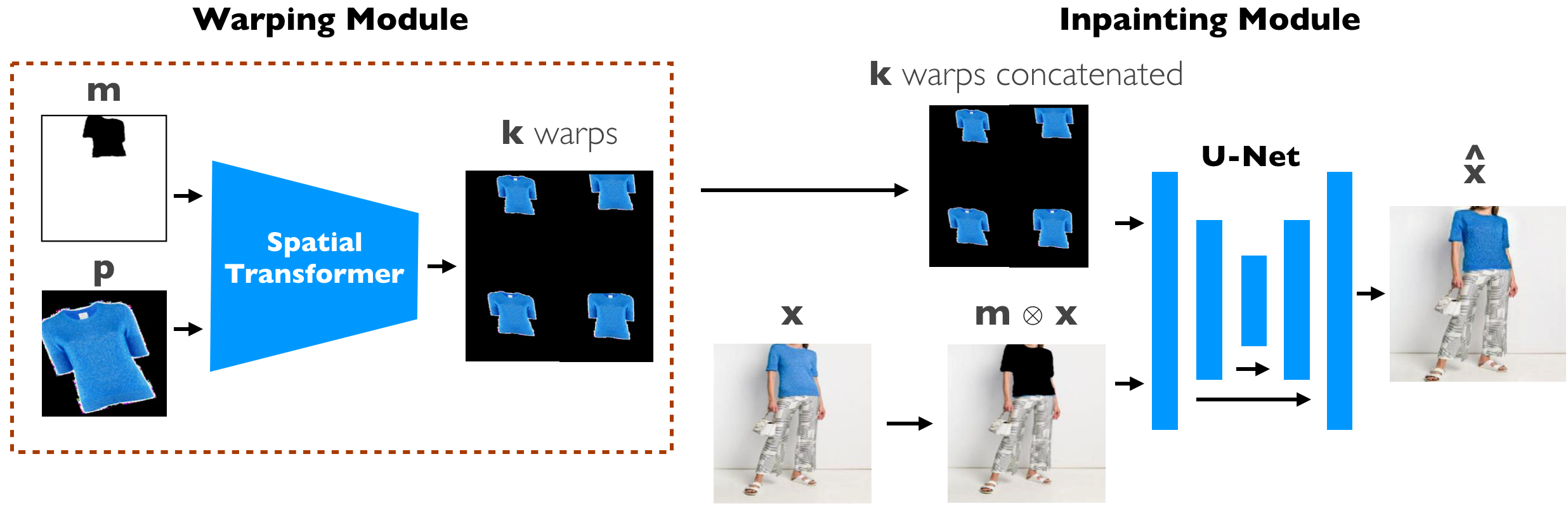}
\end{center}
   \caption{We use a warping process to place products on models. We find that using multiple specialized warpers strongly outperforms a single warper. Our warpers are trained to specialize. Having multiple warps requires the final rendering module to know which warper to rely on for different garment properties. We use a modified inpainting network that takes in the masked model ($m_i^t \otimes x_i$) and each warpers' output. This network learns to combine warps and inpaint the masked region of the model.
\vspace{-0.5cm}
}
\label{fig:synthesis-net}
\end{figure}

Our method has two components.  A {\bf Shape Matching Net} (SMN; Figure~\ref{fig:shape-embeddinng-simple} and~\ref{fig:shape-embedding}) learns an embedding for choosing shape-wise compatible garment-model pairs to perform transfer. Product and model images are matched by finding product (resp. model) images that are nearby in the embedding space.   A {\bf Multi-warp Try-on Net} (MTN; Figure~\ref{fig:synthesis-net}) takes in a garment image, a model image and a mask covering the to-change garment on the model and generates a realistic synthesis image of the model wearing the provided garment. The network consists of a warper and an inpainting network, trained jointly. The warper produces $k$ warps of the product image, each specialized on certain features. The inpainting network learns to combine warps by choosing which features to look for from each warp.
SMN and MTN are trained separately. 

%At inference time, we obtain a large collection of product images and fashion model images, and pre-compute the shape embedding for every product image and all garments on every model image (a model may wear multiple items). Given a query (product or model), we perform k-nearest neighbor search in the shape embedding to retrieve garments (product or model) that match in shape. Then, the Synthesis Net replace the fashion products on the target person, one at a time. Following this procedure, we could generate realistic studio image for an arbitrary outfit.

%Because the Synthesis Net learned to regenerate the model image from the product image and the masked model image, it can also generate reasonable synthesis using other product images of similar geometric layout. Since the product images in our dataset follow similar layout (front-view, laying flat, white background), the method generalizes well. However, using a product of different shape may yield undesirable output (as shown in figure~\ref{fig:improvement-illustration}), because the network is only trained on pairs that match on shape. 
For the rest of the paper we will define the following terms. Let $p_i^t$ represent a product image of type $t$ indexed by $i$, $x_i$ the model image, and $m_i^t$ the corresponding product mask on $x_i$. Note that $x_i$ is the groundtruth image of a model wearing product $p_i^t$. 
\subsection{Shape Matching Net}

Given an arbitrary product $p_i^t$, our goal is to retrieve a set of model images $X_i \subseteq X$ that is compatible with the shape of $p_i^t$; and vice versa.
%given an arbitrary model $o_i \in O$ and a type $t_j$, the system  can retrieve a set of product $P_i \subseteq P$ of type $t_j$ of similar shape. 
To support such query, we train a Shape Matching net that maps model and product images of similar shapes close together in an embedding space. We perform k nearest neighbors search in this embedding space to retrieve product-model pairs for creating synthesis images.

We use product images to learn a shape embedding, because product images follow a similar geometrical layout. From every product image $p_i^t$, we create a contour image $c_i^t$ by converting it into grayscale, applying a mean filter, Gaussian Adaptive Threshold and a contour finding algorithm~\cite{Suzuki1985TopologicalSA}. The contour images preserve the shape information and remove other unimportant information (e.g., color, pattern, material, etc..). A shape auto-encoder $\{E_s, D_s\}$ is trained to reconstruct the contour image $\hat{c_i}^t = D_s(E_s(c_i^t))$ using mean squared error as reconstruction loss and $l_2$ regularization on the embedding space. \begin{equation}
\mathcal{L}_{autoencoder} = \|D_s(E_s(c_i^t)) - c_i^t\|^2 + {l}_{2}
\end{equation}

When parsing a fashion model image $x_i$, we need to retrieve product information conditioned on types $t$. As our dataset contains pairs of product images and model images, we exploit such cues from the pairs and use spatial attention layers to identify the subset of features corresponding to each type of product on a model image. The garment visual encoder $E_v$ outputs an embedding vector $E_v(p_i^t)$ for a product image and an embedding vector per type $\{E_v(x_i)_{t_1}, E_v(x_i)_{t_2}, ..., E_v(x_i)_{t_n}\}$ for a model image. We embed pairs of product image $p_i^{t_a}$ of type $a$ and model image $x_i$, such that $E_v({p_i^t})$ is embedded closer to $E_v(x_i)_{t_a}$ than a different product image $E_v({p_j^t})$ or a different garment on the model $E_v(x_i)_{t_b}$ using Triplet loss $\mathcal{L}_{triplet}$~\cite{conf/cvpr/SchroffKP15}. We sample $E_v({p_j^t})$ randomly from items of the same type as $E_v({p_i^t})$ and $E_v(x_i)_{t_b}$ uniformly at random. Additionally, we minimize the squared distance between $E_v({p_i^t})$ and $E_v(x_i)_{t_a}$. An $l_2$ regularization is enforced on the embedding space.
The attention loss can be written as
\begin{multline}
    \mathcal{L}_{attention} = {\mathcal{L}_{triplet}}(E_v({p_i^t}), E_v({x_i)_{t_a}}, E_v({p_j^t})) + \\ {\mathcal{L}_{triplet}}(E_v({p_i^t}), E_v({x_i)_{t_a}}, E_v({x_i)_{t_b}}) + \|E_v({p_i^t}) - E_v(x_i)_{t_a}\|^2 + l_2
\end{multline}
The embedding loss is used to capture the feature correspondence of the two domain and help enforce the attention mechanism embed in the network architecture. Details about the spatial attention architecture are in Supplementary Materials.

To perform shape matching, we are only interested in the shape information extracted from the model image, rather than the full visual information. Therefore, we map the visual embedding $E_v(p_i^t)$ into the shape embedding $E_s(c_i^t)$ using a two-layers fully connected network $T$, such that $T(E_v(p_i^t)) = E_s(c_i^t)$. We use $D_s$ to reconstruct $c_i^t$ from $T(E_v(p_i^t))$, and compute the reconstruction loss. Additionally, we compute the triplet loss between a pair of original and transferred shape embedding, and the embedding of a different item. The loss is written as \begin{equation}
\mathcal{L}_{map} = \|D_s(T(E_v(p_i^t))) - c_i^t\|^2 + L_{triplet}(E_s(c_i^t), T(E_v(p_i^t)), E_s(c_j^t))
\end{equation}

The full training loss for the Shape Matching Net is \begin{equation}
\mathcal{L}_{matching} = \mathcal{L}_{autoencoer} + \mathcal{L}_{att} + \mathcal{L}_{map}
\end{equation}

\subsection{Multi-warp Try-on Net}

 At train time, the network takes pairs of $(p_i^t, x_i\otimes m_i^t)$ and learns to reconstruct $x_i$. At test time, ${p_i^t}$ is replaced with ${p_j^t}$ and the network generates $\hat{x}_{ij}^t$. This transfer works well when ${p_i^t}$ and ${p_j^t}$ follow similar geometric layout, ensured by the shape matching process.

As with prior work~\cite{han2017viton,wang2018toward}, our system also consists of two modules: (a) a warper to create {\bf multiple specialized warps}, by aligning the product image with the mask; (b) an inpainting module to combine the warps with the masked model and produce the synthesis image. Unlike prior work~\cite{han2017viton,wang2018toward}, the two modules are trained jointly rather than separately, so the inpainter guides the warper.

The {\bf warper} consist of a spatial transformer network~\cite{NIPS2015_5854} that takes $p_i^t$ and $m_i^t$ as input, and output $k$ sets of affine transformation parameters ${\theta_1...\theta_k}$. Then, we apply the predicted affine transformations to $p_i^t$ using ${\theta_1...\theta_k}$ to generate warps ${w_1...w_k}$. The warps are optimized to match the pixels in the masked region of the target person $(1-m_{(y,z)}) x_{(y,z)}$ using per pixel $\mathcal{L}_1$ loss written as:
\begin{equation}
\mathcal{L}_{pix}(y, z) = \\|w_{(y,z)} - (1-m_{(y,z)}) x_{(y,z)}\\|(1 + \beta (1-m_{(y,z)})).
\end{equation} $y,z$ are pixel locations; $w,h$ are the image width and height; $\beta$ controls the ratio of the loss enforced only on the mask region but not on the background. This is necessary because the majority of the masks we use are noisy, as they are produced by a pre-trained segmentation model. A balanced ratio encourage the warp to match the pixel values in the masked region, while attempting to keep all pixels within the mask region (examples in Supplementary Materials). This loss is sufficient to train a single warp baseline model.

{\bf Cascade Loss:} With multiple warps, each warp $w_i$ is trained to address the mistakes made by previous warps $w_j$ where $j < i$. For the $k$ th warp, we compute the minimum loss among all the previous warps at every pixel, written as
\begin{equation}
\mathcal{L}_{warp}(k) = \frac{\sum_{y=1, z=1}^{w,h}{ min(\mathcal{L}_{pix}(y, z)_1 \dots \mathcal{L}_{pix}(y, z)_k )}}{wh}.
\end{equation} The cascade loss computes the average loss for all warps. An additional regularization terms is enforced on the transformation parameters, so all the later warps stay close to the first warp.
\begin{equation}
\mathcal{L}_{cascade}(k) = \frac{\sum_{i=1}^k{\mathcal{L}_{warp}(i)}}{k} + \alpha\frac{\sum_{i=2}^k{\|\theta_k - \theta_1\|^2}}{k-1}
\end{equation}
The cascade loss enforce a hierarchy among all warps, making it more costly for an earlier warp to make a mistake than for a later warp. This prevents possible oscillation during the training (multiple warps compete for optimal). The idea is comparable with boosting, but yet different because all the warps share gradient, making it possible for earlier warps to adjust according to later warps.

The {\bf Inpainting Module} concatenates all the warps ($w_1...w_k$) and the masked target image ($x_i \otimes m_i^t$), and learns to inpaint the masked region on the target image. This is different from a standard inpainting task because the exact content to the masked region has been provided through warps. Rather, the Inpainting Module learns to combine the different warps to synthesize a final realistic and accurate image. We use a U-Net architecture with skip connections to help learn the identity and adopt the inpainting losses $\mathcal{L}_{inpaint}$ proposed by Liu \etal~\cite{Liu_2018_ECCV}. We also experimented adding adversarial loss and conditional adversarial loss during training, and both yield no improvement.

The total loss for the Multi-warp Try-on Net is written as
\begin{equation}
\mathcal{L}_{multiwarp}(k) = \mathcal{L}_{cascade}(k) + \mathcal{L}_{inpaint}
\end{equation}

\section{Experiments}

\subsection{Datasets}

The VITON dataset~\cite{han2017viton} contains pairs of product image (front-view, laying flat, white background) and studio images, 2D pose maps and pose key-points. It has been used by many works~\cite{wang2018toward,Dong2018SoftGatedWF,Han_2019_ICCV,YuVTNFPAI,Jandial2020SieveNetAU,Lee_2019_ICCV_Workshops,Ayush_2019_ICCV_Workshops,Raffiee2020GarmentGANPA}. Some works~\cite{Wu2018M2ETryON,Han_2019_ICCV,Grigorev2019CoordinateBasedTI,8836494} on multi-pose matching used DeepFashion~\cite{liuLQWTcvpr16DeepFashion} or MVC~\cite{Liu2016MVCAD} and other self-collected datasets~\cite{Dong2019TowardsMG,Hsieh2019FashionOnSI,Wu2018M2ETryON,arbitrary-pose}. These datasets have the same product worn by multiple people, but do not have a product image, therefore not suitable for our task. 

The VITON dataset only has tops. This likely biases performance up, because (for example): the drape of trousers is different from the drape of tops; some garments (robes, jackets, etc.) are often unzipped and open, creating warping issues; the drape of skirts is highly variable, and depends on details like pleating, the orientation of fabric grain and so on.
To emphasize these real-world problem, we collected a new dataset of 422,756 fashion products through web-scraping fashion e-commerce sites. Each product contains a product image (front-view, laying flat, white background), a model image (single person, mostly front-view), and other metadata. We use all categories except shoes and accessories, and group them into four types (top, bottoms, outerwear, or all-body). Type details appear in the supplementary materials.

%by parsing the product name using a fixed set of rules. Tops are above waist and cannot be unbuttoned or unzipped (e.g., shirt, t-shirt, sweater, tank, etc..); bottoms are below waist (e.g., pants, short, skirt, kilt, etc..); all-body cross upper and lower body (e.g., dress, gown, suit, etc..); outerwear can cover other garment and can be unzipped or unbuttoned (robe, jacket, coat, blazer, etc..).

We randomly split the data into 80\% for training and 20\% for testing. Because the dataset does not come with segmentation annotation, we use Deeplab v3~\cite{Chen2018EncoderDecoderWA} pre-trained on ModaNet dataset~\cite{zheng/2018acmmm} to obtain the segmentation masks for model images. A large portion of the segmentation masks are noisy, which further increases the difficulty (see Supplementary Materials).

\subsection{Training Process}
We train our model on our newly collected dataset and the VITON dataset~\cite{han2017viton} to facilitate comparison with prior work. When training our method on VITON dataset, we only extract the part of the 2D pose map that corresponds to the product to obtain segmentation mask, and discard the rest. The details of the training procedure is in Supplementary Materials.

We also attempted to train prior works on our dataset. However, prior work~\cite{wang2018toward,han2017viton,Dong2018SoftGatedWF,Han_2019_ICCV,YuVTNFPAI,Jandial2020SieveNetAU,Lee_2019_ICCV_Workshops,Grigorev2019CoordinateBasedTI,Wu2018M2ETryON,8836494,Chen_2019_ICCV_Workshops,Raffiee2020GarmentGANPA} require pose estimation annotations which is not available in our dataset. Thus, we only compare with prior work on the VITON dataset. 

\subsection{Quantitative Evaluation}

Quantitative comparison with state of the art is difficult.  Reporting the FID in other papers is meaningless, because the value is biased and the bias depends on the parameters of the network used~\cite{chong2019effectively,Raffiee2020GarmentGANPA}. 
We use the {\bf  FID$_\infty$} score, which is unbiased.
We cannot compute FID$_\infty$ for most other methods, because results have not been released; in fact, recent methods (eg~\cite{Han_2019_ICCV,YuVTNFPAI,Jandial2020SieveNetAU,Jandial2020SieveNetAU,Song2019SPVITONSI,Lee_2019_ICCV_Workshops,Ayush_2019_ICCV_Workshops}) have not released an implementation. 
CP-VTON~\cite{wang2018toward} has, and we use this as a point of comparison.

\begin{figure}
\begin{center}
   \includegraphics[width=0.8\linewidth]{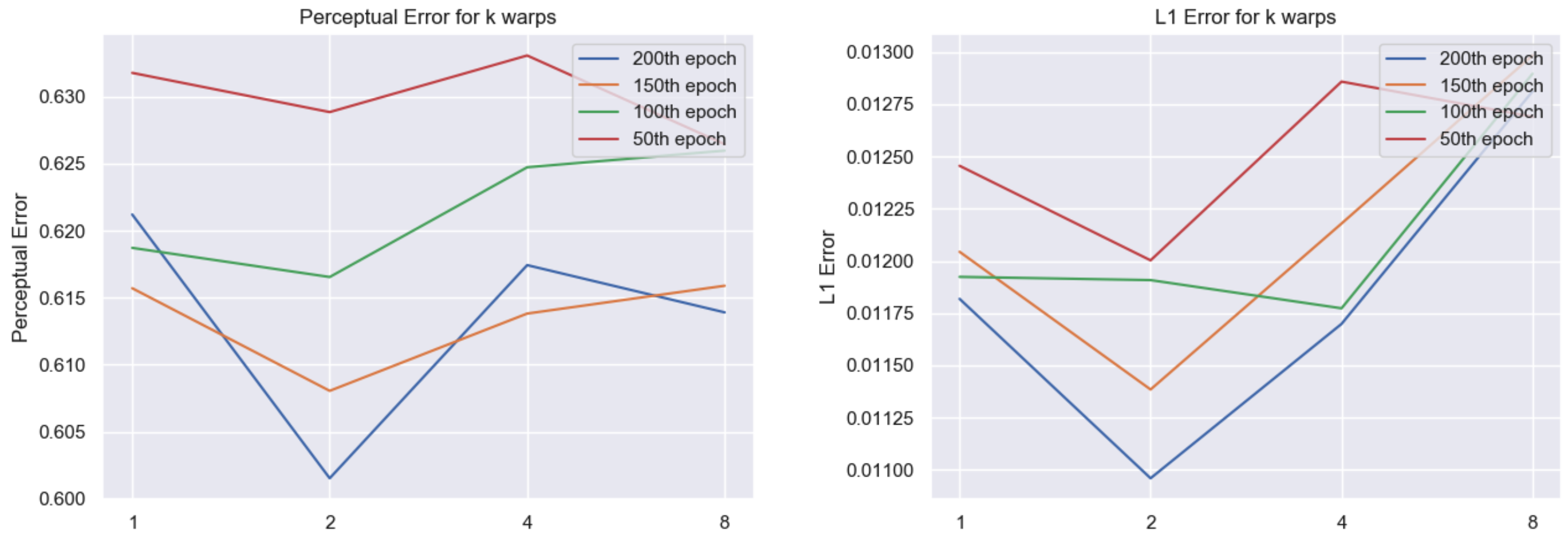}
\end{center}
   \caption{The figure compares the $\mathcal{L}1$ loss and perceptual loss (pre-trained VGG19) on the test set across 200 training epochs, recorded every 5 epochs. k=2 has the lowest error overall. Using a large $k$ speeds up the training at early stage but later overfits.
\vspace{-0.5cm}
}
\label{fig:comparison-K}
\end{figure}

Most evaluation is qualitative, and others~\cite{Jandial2020SieveNetAU,Raffiee2020GarmentGANPA} also computed the FID score on the original test set of VITON, which consists of only 2,032 synthesized pairs. Because of the small dataset, this FID score is not meaningful. The variance arising from the calculation will be high which leads to a large bias in the FID score, rendering it inaccurate. To ensure an accurate comparison, we created a larger test set of synthesized 50,000 pairs through random matching, following the procedure of the original work~\cite{han2017viton}. We created new test sets using our shape matching model by selecting the top 25 nearest neighbors in the shape embedding space for every item in the original test set. We produce two datasets each of 50,000 pairs using colored image and grayscale images to compute the shape embedding. The grayscale ablation tells us whether the shape embedding looks at color features. 

The number of warps is chosen by computing the $\mathcal{L}_{1}$ error and Perceptual error (using VGG19 pre-trained on ImageNet) using warpers with different $k$ on the test set of our dataset. Here the warper is {\em evaluated}  by mapping a product to a model wearing that product.  
As shown in figure~\ref{fig:comparison-K}, $k=2$ consistently outperforms $k=1$. However, having more than two warps also reduce performance using the current training configuration, possibly due to overfitting.

We choose $\beta$ by training a single warp model with different $\beta$ values using 10\% of the dataset, then evaluating on test.  Table~\ref{table:headings} shows that a $\beta$ that is too large or two small cause the performance to drop. $\beta=3$ happens to be the best, and therefore is adopted. Qualitative comparison are available in supplementary materials.

\begin{table}
\begin{center}
\caption{$\mathcal{L}1$ error and Perceptual error for different $\beta$ for the $\mathcal{L}_{pix}$ on test set. 
%A too large or too small $\beta$ value leads to drop in performance. 
$\beta=3$ has the best performance among the values compared.
\vspace{-0.2cm}}
\label{table:headings}
\begin{tabular}{c|cccc}
\hline
$\beta$ in $\mathcal{L}_{pix}$ & 0 & 3& 10 & 50\\
\hline
$\mathcal{L}1$ Test Error & 0.020 & { \bf0.017}& 0.019 &0.022 \\
Perceptual Test Error &0.774 & { \bf 0.722} & 0.745 & 0.810\\
\hline
\end{tabular}
\end{center}
\end{table}
\vspace{-0.3cm}

With this data, we can compare CP-VTON, our method using a single warp ($k=1$), and two warps ($k=2$), and two warp blended. The blended model takes in the average of two warps instead of the concatenation. 
Results appear in Table~\ref{table:fid}. We find:
\begin{itemize}
\item for all methods, choosing the model gets better results;
\item there is little to choose between color and grayscale matching, so the match attends mainly to garment shape;
\item having two warpers is better than having one;
\item combining with a u-net is much better than blending.
\end{itemize}
We believe that quantitative results understate the improvement of using more warpers, because the quantitative measure is relatively
crude. Qualitative evidence supports this (figure~\ref{fig:multi-warp-qualitative}).

\begin{table}[]
\begin{center}
\label{table:fid}
\begin{tabular}{c|c|c|c}
\hline
Test set           & Random & Match (color) & Match (grayscale) \\ \hline
CP-VTON            & 15.29  & 13.69         & 13.69             \\ \hline
Ours k=1           & 10.52  & 7.22          & 7.16              \\ 
Ours k=2           & 9.89   & \textbf{7.04} & 7.06              \\ 
Ours k=2 (blended) & 15.4   & 15.26         & 15.37             \\ \hline
\end{tabular}
\end{center}
\caption{This table compares the FID$_\infty$ score (smaller better) between different image synthesis methods on random pairs vs. matching pairs using our shape embedding network. All values in col. 1 are significantly greater than that of col. 2 and 3, demonstrating choosing a compatible pair significantly improves the performance of our methods and of CP-VTON.  We believe this improvement applies to other methods, but
others have not published code. Across methods, our method with two warpers significantly outperforms prior work on all test sets. There is not much to choose between color and grayscale matcher,
suggesting that the matching process focuses on garment shape (as it is trained to do).  Using two warps ($k=2$) shows slight improvement from using a single warp ($k=1$), because the improvements are difficult for any  quantitative metrics to capture. The difference is more visible in qualitative examples (figure~\ref{fig:multi-warp-qualitative}).
It is important to use a u-net to combine warps; merely blending produces poor results (last row).
\vspace{-0.5cm}}  
\end{table}

\subsection{Qualitative Results}
\begin{figure}
\begin{center}
   \includegraphics[width=1\linewidth]{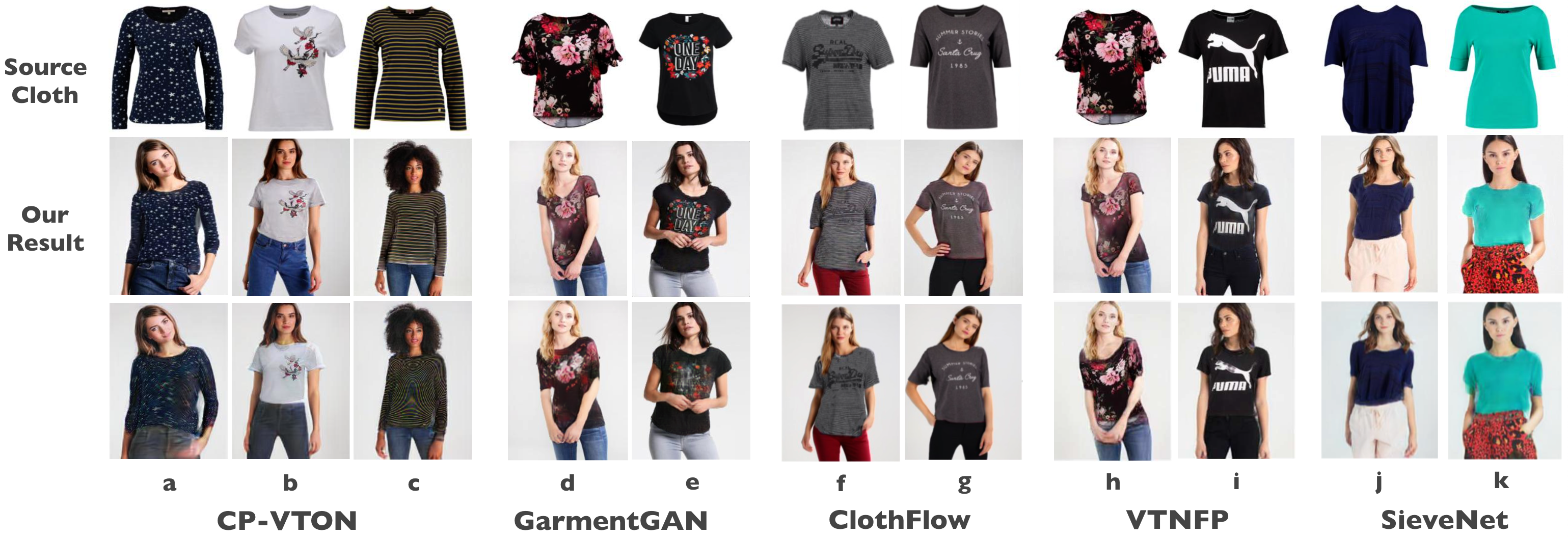}
\end{center}
   \caption{Comparisons to CP\_VTON, ClothFlow, VTNFP and SieveNet on the VITON dataset, using images published 
for those methods. Each block shows a different dataset.  Our results are in row 2, and comparison method results are in row 3. 
Note CP-VTON, in comparison to our method: obscuring necklines (b); aliasing stripes (c); rescaling transfers (b); smearing texture and blurring boundaries (a); and blurring transfers (b). 
Note GarmentGAN, in comparison to our method: mangling limb boundary (d); losing contrast on flowers at waist (d); and aliasing severely on a transfer (e).
Note ClothFlow, in comparison to our method: NOT aliasing stripes (f); blurring hands (f, g);
blurring anatomy (clavicle and neck tendons, g); rescaling a transfer (g).  
Note VTNFP, in comparison to our method:  misplacing texture detail (blossoms at neckline and shoulder, h);
mangling transfers (i).  
Note SieveNet, in comparison to our method:  blurring outlines (j, k); misplacing cuffs (k);
mangling shading (arm on k).  {\em Best viewed in color at high resolution.}
\vspace{-0.5cm}  
%   The {\bf right figure} 
   }
\label{fig:comparison-VITON}
\end{figure}

We have looked carefully for matching examples in~\cite{Han_2019_ICCV,Jandial2020SieveNetAU,YuVTNFPAI,Raffiee2020GarmentGANPA} to produce qualitative comparisons.
Comparison against MG-VTON~\cite{Dong2019TowardsMG} is not applicable, as the work did not include any fixed-pose qualitative example. 
Note that the comparison favors prior work because our model trains and tests only using the region corresponding to the garment in the 2D pose 
map while prior work uses the full 2D pose map and key-point pose annotations. 

Generally, garment transfer is hard, but modern methods now mainly fail on details.  This means that evaluating transfer requires careful attention to detail. 
Figure~\ref{fig:comparison-VITON} shows some
comparisons.  In particular, attending to image detail around boundaries, textures, and garment details exposes some of the difficulties in the task.
As shown in Figure~\ref{fig:comparison-VITON} left, our method can handle complicated texture robustly (col. a, c) and preserve details of the logo accurately (col. b, e, f, g, i). The examples also show clear difference between our inpainting-based method and prior work -- our method only modifies the area where the original cloth is presented. This property allows us to preserve the details of the limb (col. a, d, f, g, h, j) and other clothing items (col. a, b) better than most prior work. Some of our results (col. c, g) show color artifacts from the original cloth on the boundary, because the edge of the pose map is slightly misaligned (imperfect segmentation mask). This confirms that our method rely on fine-grain segmentation mask to produce high quality result. Some pairs are slightly mis-matched in shape(col. d, h). This will rarely occur with our method if the test set is constructed using the shape embedding. Therefore, our method does not attempt to address it.

\begin{figure}
\begin{center}
   \includegraphics[width=0.8\linewidth]{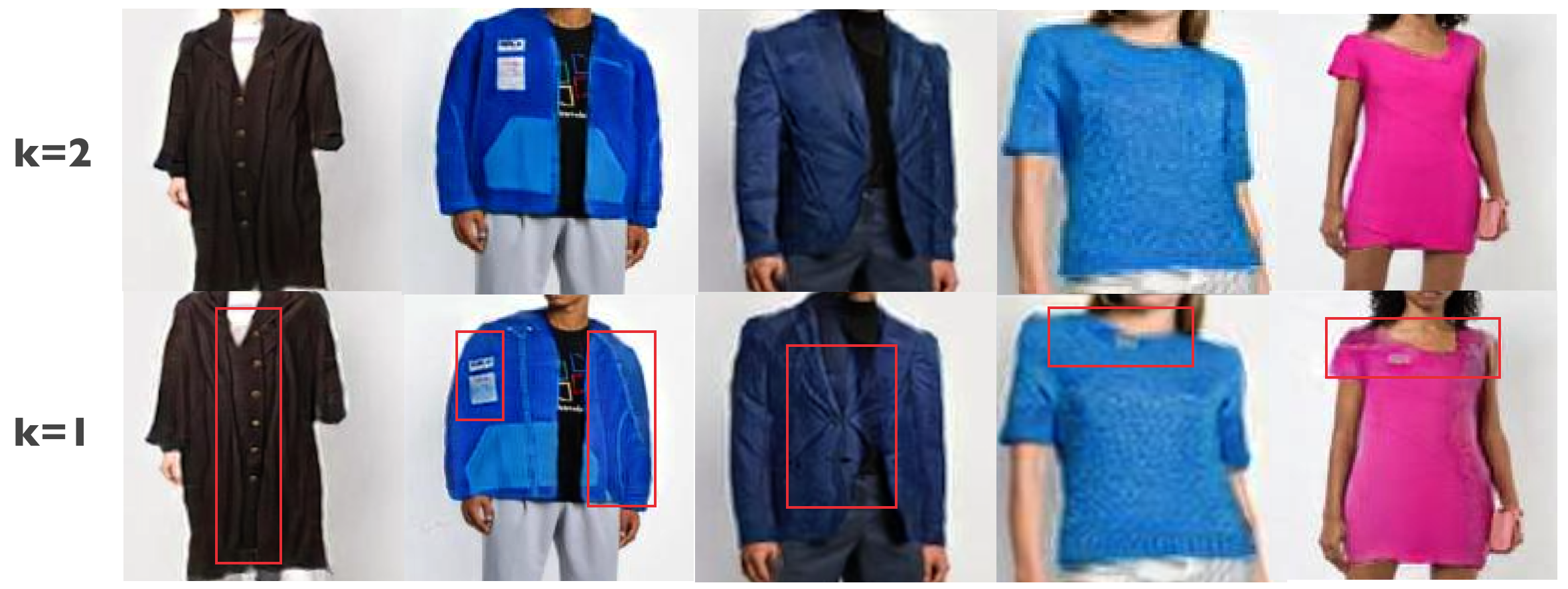}
\end{center}
   \caption{ The figures shows qualitative comparison between $k=2$ and $k=1$.   Note: the buttons in the wrong place for a single warp on the {\bf left}, fixed for $k=2$;
a misscaled pocket and problems with sleeve boundaries for the single warp on the {\bf center left}, fixed for $k=2$; a severely misplaced button and surrounding buckling in the {\bf center}, fixed for $k=2$;
a misplaced garment label on the {\bf center right}, fixed for $k=2$; another misplaced garment label on the {\bf right}, fixed for $k=2$.
   \vspace{-0.5cm}
}
   \label{fig:multi-warp-qualitative}
\end{figure}

Two warps are very clearly better than one (Figure~\ref{fig:multi-warp-qualitative}), likely because the second warp can fix the alignment and details that single warp model fails to address. 
Particular improvements occur for unbuttoned/unzipped outerwear and for product images with tags. These improvement may not be easily captured by quantitative evaluation
because the differences in pixel values are small. 

We attempted to train the geometric matching module (using TPS transform) to create warps on our dataset, as it was frequently adopted by prior work~\cite{han2017viton,wang2018toward,Dong2018SoftGatedWF}. However, TPS transform failed to adapt to partitions and significant occlusions (examples in Supplementary Materials).

\subsection{User Study}

\begin{figure}
\begin{center}
   \includegraphics[width=0.7\linewidth]{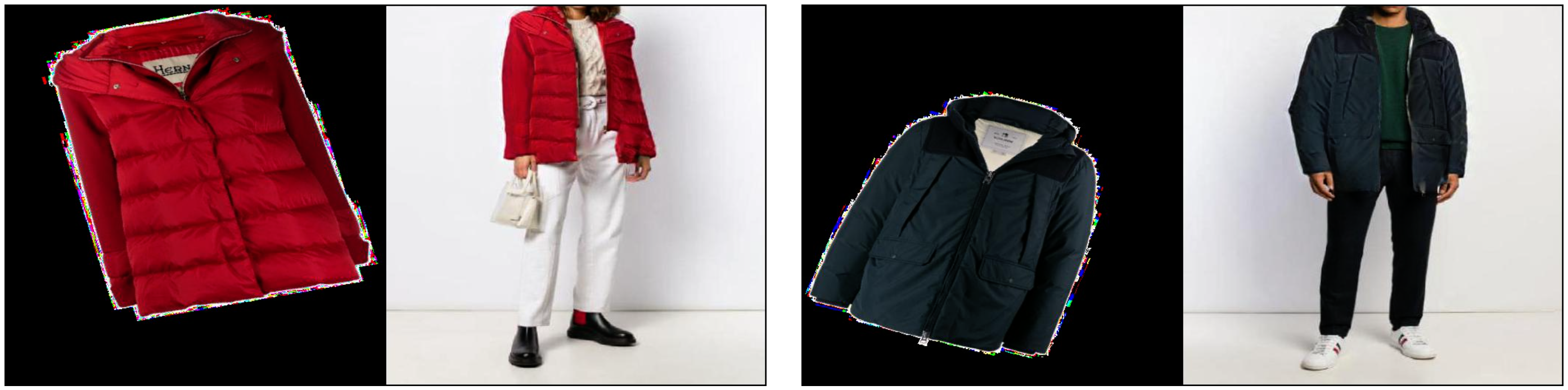}
\end{center}
   \caption{ Two synthesized images that 70\% of the participants in the user study thought were real. 
Note, e.g., the shading, the wrinkles, even the zip and the collar.
   }
\label{fig:user-study-examples}
\end{figure}

%\begin{table}
%\begin{center}
%\caption{The user study results shows that participants have high difficulties distinguish between real and synthesized images. 51.6\% and 61.5\% of the fake image are thought to be real by crowds and researchers, respectively. Occasionally, some of the real image are also thought as fake, suggesting that participants paid attention.
%\vspace{-0.2cm}}
%\label{table:user-study}
%\begin{tabular}{l|l|lll}
%\hline
% &Participants & Accuracy & False Positive & False Negative\\
%\hline
%General Population & 31 & 0.573 & { \bf 0.516}& 0.284 \\
%Vision Researcher & 19 &0.655 & { \bf 0.615} & 0.175\\
%\hline
%\end{tabular}
%\end{center}
%\end{table}

We used a user study to check how often users could identify synthesized images. A user is asked whether an image of a model wearing a product (which is shown) is real or synthesized.
Display uses the highest possible resolution (512x512), as in figure~\ref{fig:user-study-examples}.

\begin{table}
\begin{center}
\caption{The user study results show that participants have high difficulties distinguish between real and synthesized images. 51.6\% and 61.5\% of the fake image are thought to be real by crowds and researchers, respectively. Occasionally, some of the real image are also thought as fake, suggesting that participants paid attention.
\vspace{-0.2cm}}
\label{table:user-study}
\begin{tabular}{l|l|lll}
\hline
 &Participants & Accuracy & False Positive & False Negative\\
\hline
General Population & 31 & 0.573 & { \bf 0.516}& 0.284 \\
Vision Researcher & 19 &0.655 & { \bf 0.615} & 0.175\\
\hline
\end{tabular}
\end{center}
\end{table}

We used examples where the mask is good, giving a fair representation of the top 20 percentile of our results. 
Users are primed with two real vs. fake pairs before the study. Each participant is then tested with 50 pairs of 25 real and 25 fake, without repeating products. We test two populations of users (vision researchers, and randomly selected participants).

Mostly, users are fooled by our images; there is a very high false-positive (i.e. synthesized image marked real by a user) rate (table~\ref{table:user-study}). Figure~\ref{fig:user-study-examples} shows two examples of synthesized images that 70\% of the general population reported as real. They are hard outerwear examples with region partition and complex shading. Nevertheless, our method managed to generate high quality synthesis. See supplementary material for all questions and complete results of the user study.

\section{Conclusions}
In this paper, we propose two general modifications to the virtual try-on framework: (a) carefully choose the product-model pair for transfer using a shape embedding and (b) combine multiple coordinated warps using inpainting. Our results show that both modifications lead to significant improvement in generation quality. Qualitative examples demonstrate our ability to accurately preserve details of garments. This lead to difficulties for shoppers to distinguish between real and synthesized model images, shown by user study results.

\clearpage
\bibliographystyle{splncs04}
\bibliography{egbib}

\end{document}